\documentclass[11pt]{article}
\usepackage{nodalida2021}
\usepackage{times}
\usepackage{url}
\usepackage{latexsym}
\usepackage{hyperref}
\aclfinalcopy 
\def\aclpaperid{88} 

\usepackage{times}
\usepackage{latexsym}

\usepackage{caption}
\usepackage{subcaption}

\usepackage{url}
\usepackage{graphicx}
\usepackage{mathtools}
\usepackage{booktabs}
\usepackage{multirow}
\usepackage{caption,subcaption}
\usepackage{comment}
\usepackage[colorinlistoftodos]{todonotes}
\usepackage{enumerate}
\usepackage[shortlabels]{enumitem}
\usepackage{tikz}

\newcommand{\COMMENT}[1]{}

\title{Understanding Cross-Lingual Syntactic Transfer \\ in Multilingual Recurrent Neural Networks}

\author{Prajit Dhar \qquad   Arianna Bisazza  \\
 Center for Language and Cognition \\
University of Groningen  \\ 
    {\tt \{p.dhar, a.bisazza\}@rug.nl} 
\\}

\date{}

\begin{document}
\maketitle
\begin{abstract}
It is now established that modern neural language models can be successfully trained on multiple languages simultaneously without changes to the underlying architecture. 
But what kind of knowledge is really shared among languages within these models? Does multilingual training mostly lead to an alignment of the lexical representation spaces or does it also enable the sharing of purely grammatical knowledge? 
In this paper we dissect different forms of cross-lingual transfer and look for its most determining factors,
using a variety of models and probing tasks.
We find that exposing our LMs to a related language does not always increase grammatical knowledge in the target language, and that optimal conditions for \textit{lexical-semantic} transfer may not be optimal for \textit{syntactic} transfer.
\end{abstract}


\section{Introduction}



\begin{figure}[h]
\centering
\includegraphics[width=.75\linewidth]{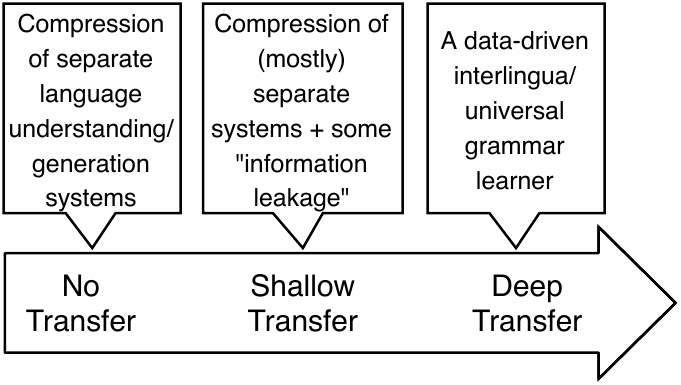}
\caption{Possible characterizations of a trained mNN in terms of cross-lingual transfer levels.}
\label{types}
\end{figure}

One of the most important NLP discoveries of the past few years has been that a \textit{single} neural network can be successfully trained to perform a given NLP task in \textit{multiple} languages without architectural changes compared to monolingual models \cite{Ostling2017multilm,Johnson16}.
Besides important practical advantages (fewer parameters and models to maintain),
such multilingual Neural Networks \mbox{(mNNs)} provide an easy but powerful way to transfer task-specific knowledge from high- to low-resource languages  \cite{devlin2018bert,lample2019cross,Aharoni,neubig-hu-2018-rapid,arivazhagan,Artetxe,chi2020finding}.
%
These success stories have led to a need for understanding  \textit{how} exactly cross-lingual transfer works within these models. 
Figure \ref{types} illustrates different possible characterizations of a trained \mbox{mNN}:
While the \mbox{no-transfer} scenario is rather easy to rule out, understanding which linguistic categories are shared, and to what extent, is more challenging.

In this work, we focus on the transfer of \textit{syntactic} knowledge among languages and look for evidence that mNNs induce a shared syntactic representation space \textit{while not receiving any direct cross-lingual supervision}.
To be clear, if we measure transfer among languages X and Y, every training sentence for language modeling will be either in language X or Y, while for machine translation every sentence pair will be either in language pair (X, Z) or (Y, Z). 
Thus, the only pressure to share linguistic representations is given by the sharing of the hidden layer parameters (as well as, possibly, some of the word embeddings).

Neural language models have been shown to implicitly capture non-trivial structure-sensitive phenomena like long-range number agreement
\cite{linzen2016,Gulordava2018,marvin-linzen-2018-targeted}. However most of these studies have been confined to monolingual models. 
%
We then investigate the following questions:
\begin{enumerate}
    \item 
    Does mNNs' implicit syntactic knowledge of L2 increase by exposure to a related L1? 
    \item 
    Do mNNs induce a common representation space with shared syntactic categories?
\end{enumerate}
Our research questions are reminiscent of well-known questions in the fields of psycholinguistic and second language acquisition, where work has shown that both lexical and syntactic representations are shared in the mind of bilinguals \cite{hartsuiker2004,vasilyeva2010}.
Taking inspiration from this body of work, we investigate what factors are needed for mNNs to successfully transfer linguistic knowledge, including vocabulary overlap, language relatedness, number of training languages, training regime (joint \textit{vs} sequential) and training objective (next word prediction \textit{vs} translation to a third language).

In contrast to the current mainstream focus on BERT-like models \cite{rogers20bertology}, we evaluate more classical LSTM-based models trained for next word prediction or translation over a moderate number of languages (2 or 9).
We choose this setup because (i) it allows for more controlled and easy-to-replicate experiments in terms of both training data and model configuration and (ii) LSTMs trained on a standard sequence prediction objective are more cognitively plausible and directly applicable to our main probing task, namely agreement prediction.
In this setup, we find limited and rather inconsistent evidence for the transfer of implicit grammatical knowledge when semantic cues are removed \cite{Gulordava2018}.
While moderate PoS category transfer occurs, truly language-agnostic syntactic categories (such as \textit{noun} or \textit{subject}) do not seem to emerge in our mNN representations. 
Finally, we find that optimal conditions for lexical-semantic transfer may not be optimal for syntactic transfer.



\section{Previous Work}

\paragraph{Multilingual Machine Translation}

Early work on multilingual NMT focused on building dedicated architectures \cite{dong2015,FiratCB16,Johnson16}. Starting from \cite{Johnson16}, m-NMT models are mostly built with the same architecture as their monolingual counterparts, by simply adding language identifying tags to the training sentences.
Using a small set of English sentences and their Japanese and Korean translations, \newcite{Johnson16} showed that semantically equivalent sentences form well-defined clusters in the high-dimensional space induced by a NMT encoder trained on large-scale proprietary datasets.
\newcite{kudugunta-19} analyze the similarity of encoder representations of different languages within a massively m-NMT model. They find that representation similarity correlates strongly with linguistic similarity and that encoder representations diverge based on the target language. However they do not disentangle the syntactic aspect from other types of transfer.



\paragraph{Multilingual Sentence Encoders}
A related line of work focuses on mapping sentences from different languages into a common representation space to be used as features in downstream tasks where training data is only available in a different language than the test language. 
\newcite{Artetxe} use the encoder representations produced by a massively multilingual NMT system similar to \cite{Johnson16} to perform cross-lingual textual entailment (XNLI) and document classification. 
m-BERT \cite{devlin2018bert,mBERT} and XLM \cite{lample2019cross} are large-scale mNNs trained on a masked LM (MLM) objective using mixed-language corpora. This results in general-purpose contextualized word representations that are multilingual in nature, \textit{without} requiring any parallel data. 
m-BERT representations have been proved particularly successful for transferring dependency parsers to low- (or zero-)resource languages \cite{wu-dredze-2019-beto,kondratyuk,tran-bisazza-19}.
On the task of cross-lingual textual entailment \cite{conneau2018xnli}, XLM-based classifiers come surprisingly close to systems that use fully-supervised MT as part of their pipeline (to translate the training or test data). 


\paragraph{Implicit Learning of Linguistic Structure}
NNs trained for downstream tasks such as language modeling, translation or textual entailment, have been shown to implicitly encode a great deal of linguistic structure such as morphological features \cite{Belinkov,lazyencoder,bjerva-augenstein-2018-phonology}, number agreement \cite{linzen2016,Gulordava2018} and other structure-sensitive phenomena \cite{marvin-linzen-2018-targeted}. Studies such as \cite{tenney19-iclr,tenney19-acl,jawahar19} have extended these findings to BERT representations showing positive results on a variety of syntactic probing tasks.
Extensive overviews of this body of work are presented in \cite{belinkov-survey} and \cite{rogers20bertology}. 
%




\paragraph{Cross-lingual Transfer in Multilingual NNs}
Recent studies \cite{wu-dredze-2019-beto,howmbert,chi2020finding} have found evidence of \textit{syntactic} transfer in m-BERT using POS tagging and dependency parsing experiments.
On the other hand, \newcite{semanticbert} find that \mbox{m-BERT} representations capture cross-lingual \textit{semantic} equivalence sufficiently well to allow for word-alignment and sentence retrieval, but fail at the more difficult task of MT quality estimation.
While this massive Transformer-based \cite{Vaswani:2017} architecture has received overwhelming attention in the past year, we believe that smaller, better understood, and easier to replicate model configurations can still play an important role in the pursuit of NLP model explainability.
Moreover, the large number of m-BERT training languages (ca. 100) added to the uneven language data distribution and the highly shared subword vocabulary, make it difficult to isolate transfer effects in any given language pair.
\newcite{mueller-etal-2020-cross} recently tested a LSTM trained on five languages on a multilingual extension of the subject-verb agreement set of \newcite{marvin-linzen-2018-targeted}.
They found signs of harmful interference rather than positive transfer across languages.
In Section~\ref{sect:exposure} we corroborate this rather surprising finding by using a more favourable setup for transfer, that is: (i) only two, related, training languages, (ii) a simulated low-resource setup for the target language, and (iii) eliminating vocabulary overlap during training with language IDs. 




\paragraph{Cross-lingual Transfer in the Bilingual Mind}
Measuring the extent to which dual-language representations are shared in the mind of bilingual subjects is a long-standing problem in the field of second language acquisition \cite{Kellerman:86,Odlin:89,Jarvis:08,Kootstra:12}.
Among others, \newcite{Hartsuiker:04} present evidence of cross-lingual \textit{syntactic priming} in bilingual English-Spanish speakers, which are more inclined to produce English passive sentences after having heard a Spanish passive sentence.
Using neuroimaging techniques in a reading comprehension experiment with in German-English bilinguals, \newcite{kristenenggerman}  report that the processing of L1 and L2 sentences activates the same brain areas, pointing to the shared nature of syntactic processing in the bilingual mind.
%
%
Taking inspiration from this body of work, we investigate what factors trigger cross-lingual transfer of syntactic knowledge within mNNs. 



\paragraph{Cross-Lingual Dependency Parsing}
Finally, our work is also related to the productive field of cross-lingual and multilingual dependency parsing \citep[\textit{inter alia}]{naseem2012selective,zhang2015hierarchical,tackstrom2012cross,ammar2016many}, with the important difference that we are interested in models that are \textit{not} explicitly trained to recognize syntactic structure but acquire it indirectly while optimizing next word prediction or translation objectives.
Among others, \newcite{ahmad-etal-2019-difficulties} have shown that the difficulty of transferring a dependency parser cross-lingually depends on typological differences between the source and target languages, with word order differences playing an important role.
In this paper, we mainly consider source-target languages that are related, like French or Spanish (source) and Italian (target), where we expect implicit syntactic knowledge to be more easily transferable.

\section{Probing Tasks}

To answer our RQ1 (are mNNs capable of implicitly transferring syntactic knowledge between languages?) we choose the task of Number Agreement. For our RQ2 (are mNNs able to induce a common representation with shared syntactic categories?) we look at less complex syntactic tasks such as PoS tag classification 
and Dependency relation classification, and contrast them with a lexical-semantic task (word translation retrieval).
We choose these tasks because they can be framed as simple classification (or ranking) problems and have a direct linguistic interpretation.
We do not consider parsing because it is a complex task with a highly structured prediction space requiring dedicated model components.
The probed models are LSTM-based language models and translation models, trained at the word-level. More details are provided below.



\subsection{Number Agreement}
\label{agree}

Number agreement describes the instance where a phrase and its arguments or modifiers must agree in their number feature. Number agreement can occur between a subject-predicate pair (\textit{the son$_{sg}$ of my neighbors goes$_{sg}$)}, noun-quantifier pair \textit{(many$_{pl}$ huge trees$_{pl}$)}, etc. 
\newcite{linzen2016} first proposed the subject-verb agreement task to assess the ability of a LSTM-based LM to capture non-trivial language structure, by checking if the correct verb form was assigned a higher probability than the wrong one, e.g. if prob(\textit{were}$|$context) $>$ prob(\textit{was}$|$context) in the sentence \textit{The boys, who were lost in the forest \textbf{were}/\textbf{was} found}.
LM performance was shown to be mostly affected by the number of agreement attractors.

\paragraph{Probing Dataset}
We adopt the benchmark by \newcite{Gulordava2018}, henceforth called G18, which extends the evaluation  of \newcite{linzen2016} to more languages and more agreement constructions, automatically harvested from corpora using POS patterns.
G18 also introduced two conditions to test whether a model relies on semantic cues or purely grammatical knowledge to predict agreement:
\begin{enumerate}
    \item Original : Sentences automatically extracted from corpora;
    \item Nonce : Nonsensical but grammatical sentences created by randomly replacing all content words in the original sentence with random words with same morphological class.
\end{enumerate}
Thus, this is one of few existing tasks that allow us to measure the transfer of grammatical knowledge in isolation. 
Using the G18 benchmark, we compare mNNs with monolingually trained models, in order to compare if the addition of a related language improves the long-range agreement accuracy of the monolingual model.
We expect this to happen for languages that have the same number agreement patterns, like French and Italian.

%

\begin{figure}
\centering
\includegraphics[width=\linewidth]{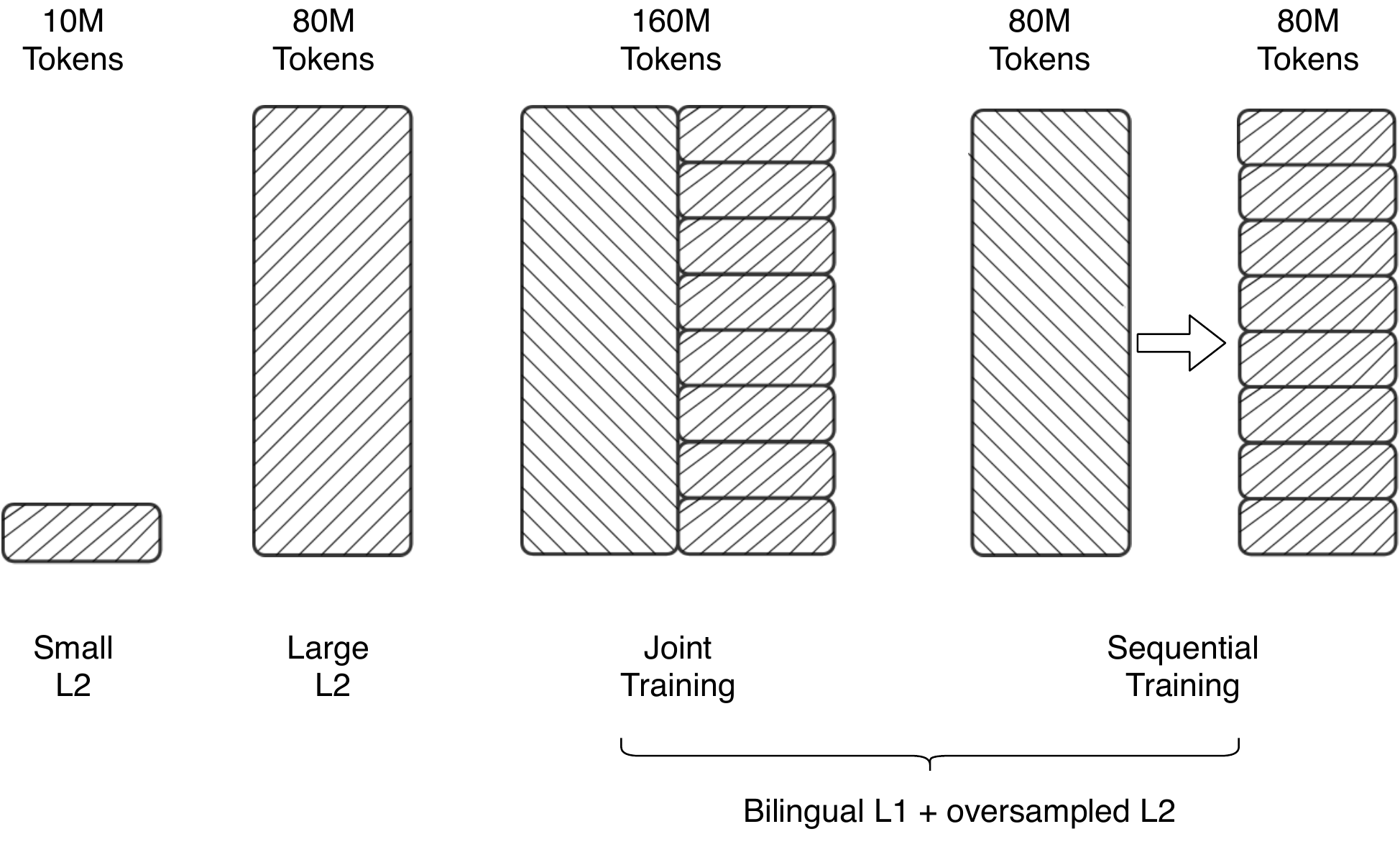}
\captionof{figure}{Monolingual and bilingual LM training schemes used in our agreement experiments.}
\label{training}
\end{figure}

\paragraph{Probed Models}
Similar to G18, we train 2-layer LSTMs with embedding and hidden layer size of 650, for 40 epochs, using a dataset of crawled Wikipedia articles. 
These language models are trained on next word prediction and do not receive any specific supervision for the syntactic task.
$L1$ is our helper language and $L2$ is the target language where we measure agreement accuracy. 
Fig.~\ref{training} shows our different training setups.
To simulate a low-resource setup and possibly increase the chances of transfer, 
we train our bilingual LMs on a shuffled mix of a larger L1 corpus (L1$_{large}$, 80M tokens) and a smaller L2 corpus ($L2_{small}$, 10M tokens).
L2 is oversampled to approximately match the amount of L1 sentences. 
This bilingual model (LM$_{L1+L2_{small}}$) is compared to a baseline monolingual LM trained on a small L2 corpus (LM$_{L2_{small}}$).
As upper bound, we also show the results of a model trained on more L2 data (80M). This model performs closely to the results reported by G18 with a similar setup.

Most experiments in this paper are performed by \textit{joint training}, i.e. the model is trained on the mixed language data since initialization.
However in Sect.~\ref{sect:trainvocab} we also evaluate \textit{pre-training}: i.e. the LM is first trained on L1 data, then after convergence, it continues training on L2 data (see Fig.~\ref{training}). 
%
A language tag is introduced at the beginning of each sentence.
The vocabulary for each language consists of the 50k most frequent tokens, with the remaining tokens replaced by the unknown tag.
The bilingual vocabulary is the union of the language-specific vocabularies, resulting in a total of 88k words in our main language pair (French-Italian).
In Sect.~\ref{sect:trainvocab} we compare this setup (called \textit{natural overlap}) to a \textit{no-overlap} setup where all words are prepended with a language tag, resulting in a bilingual vocabulary of 100k words. 
 



\subsection{Cross-lingual Syntactic Category Classification}
\label{syntacticclass}


To verify whether basic syntactic categories are shared among different language representations in mNNs, we inspect the activations of our trained LMs when processing a held-out corpus. Specifically we build linear classifiers to predict either the PoS tag or the Dependency label (type of relation to the head) of a word from its hidden layer representation. 
This setup is similar to previous work \cite{Blevins,tenney19-iclr}, however our diagnostic classifiers are trained on L1 and tested on L2.\footnote{Another difference regards the dependency classification: \newcite{Blevins} uses constituency parsing and \newcite{tenney19-iclr} predicts dependency arcs given word \textit{pairs}.} 
If syntactic categories are shared, we expect to see minor drops in classification accuracy compared to a classifier trained and tested on L2.
In other words, we ask whether, e.g., French and Italian adjectives or subjects are recognizable by the same NN activations.

Several studies such as \cite{lazyencoder,hewitt-liang-2019-designing,pimentel-etal-2020-information} have criticised diagnostic classifiers for overestimating the ability of neural networks to capture linguistic information. We partly address these pitfalls by comparing classification accuracy on top of our trained mNNs with that of their corresponding randomly initialized counterparts.

\paragraph{Probing Dataset}
We probe our models on manually annotated coarse-grained PoS and Dependency labels taken from Universal Dependency Treebanks \cite{ud2.4}.
Specifically, we use French-GSD (389k tokens), Italian-ISDT (278k), 
Spanish-AnCora (548k), and German-GSD (288k). 
UD sentences are fed to a trained model's encoder and the resulting last-layer activations are used to build the probing classifiers.

\paragraph{Probed Models}
We first apply the PoS and Dependency probing tasks to the Wikipedia-based LMs described in Sect.~\ref{agree}.
To study the effect of training objective (next word prediction \textit{vs} translation to a third language), in Sect.~\ref{sect:rq2europarl} we perform another set of controlled experiments using the Europarl\footnote{http://www.statmt.org/europarl/v7/} parallel corpus. Our dataset consists of $L1$ $\rightarrow$ English parallel sentences, where $L1$ is one of nine languages chosen from three different families: French, Italian, Portuguese, Spanish (Romance); German, Dutch, Swedish and Danish (Germanic) and Finnish (Uralic), with about 45.9M tokens for each language pair.
%
The NMT models implement a standard attentional sequence-to-sequence architecture based on  \mbox{4-layer} bidirectional LSTMs \cite{bahdanau14} with embedding and hidden layer size of 1024. 
%
To maximize comparability between translation and language modeling objectives, the LMs in these experiments are also 4-layer bidirectional (BiLMs, \`{a} la \newcite{elmo}) with the same hidden layer size, trained on the source-side portion of our Europarl dataset.

\subsection{Word Translation Retrieval}
\label{word}
To put syntactic transfer in contrast with other types of transfer effects, we also experiment with word translation retrieval (henceforth abbreviated as WTR).
This was used as a probing task for cross-lingual word embeddings in \cite{lample2017unsupervised,conneau2017word} and involves calculating the distance (measured by cosine similarity) between the embedding of a source language word (e.g., \textit{bonjour}) and that of its translation (e.g., \textit{buongiorno}).
Since the task is context independent, only the word-type embeddings are probed.
We interpret precision in this task as a measure of the alignment of two word embedding spaces, that is \textit{lexical-semantic} transfer.



\paragraph{Lexicon}
The bilingual lexicon from MUSE \cite{lample2017unsupervised} is used as gold standard for this task. MUSE is available for several language pairs and includes polysemous words (many-to-many pairs). For each language pair, we use 1.5k source and 200k target words.



\section{Does Exposure to L1 Improve Implicit Syntactic Knowledge on a Related L2?}
\label{sect:exposure}
To answer RQ1 we use the number agreement task, which is explained in detail in Sect. \ref{agree}. 
We choose Italian (IT) and Russian (RU) from the G18 dataset as our target languages $L2$. As helper languages, $L1$, we choose French (FR) and Spanish (ES) for $L2$ IT, and FR and Ukrainian (UK) for $L2$ RU, which allows us to study the impact of language relatedness.
%
Accuracy is calculated as follows: for each sentence in the $L2$ benchmark, if the probability of the correct verb form is higher than the incorrect form, the agreement is said to be correct, and incorrect otherwise. 

\begin{figure*}[ht]
\centering
\begin{subfigure}{0.48\textwidth}
\caption{Agreement accuracy on L2: Italian}
\includegraphics[width=\textwidth]{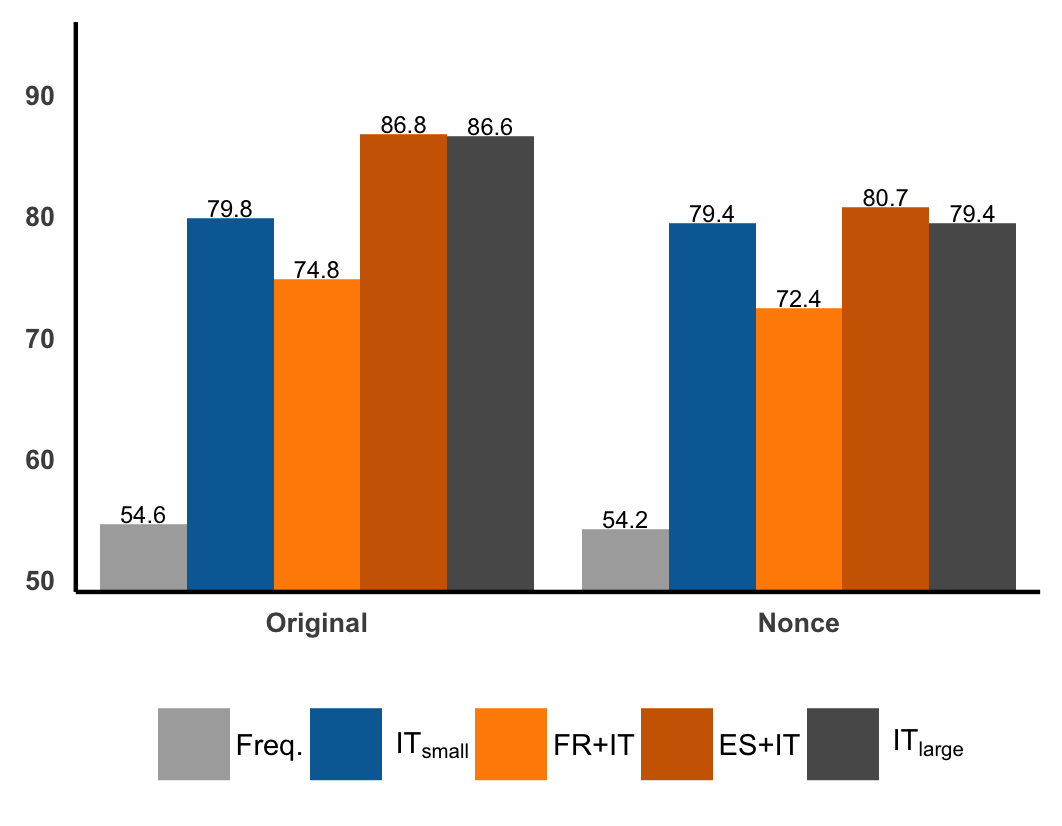}
\label{fig:it}
\end{subfigure}
\begin{subfigure}{0.48\textwidth}
\caption{Agreement accuracy on L2: Russian}
\includegraphics[width=\textwidth]{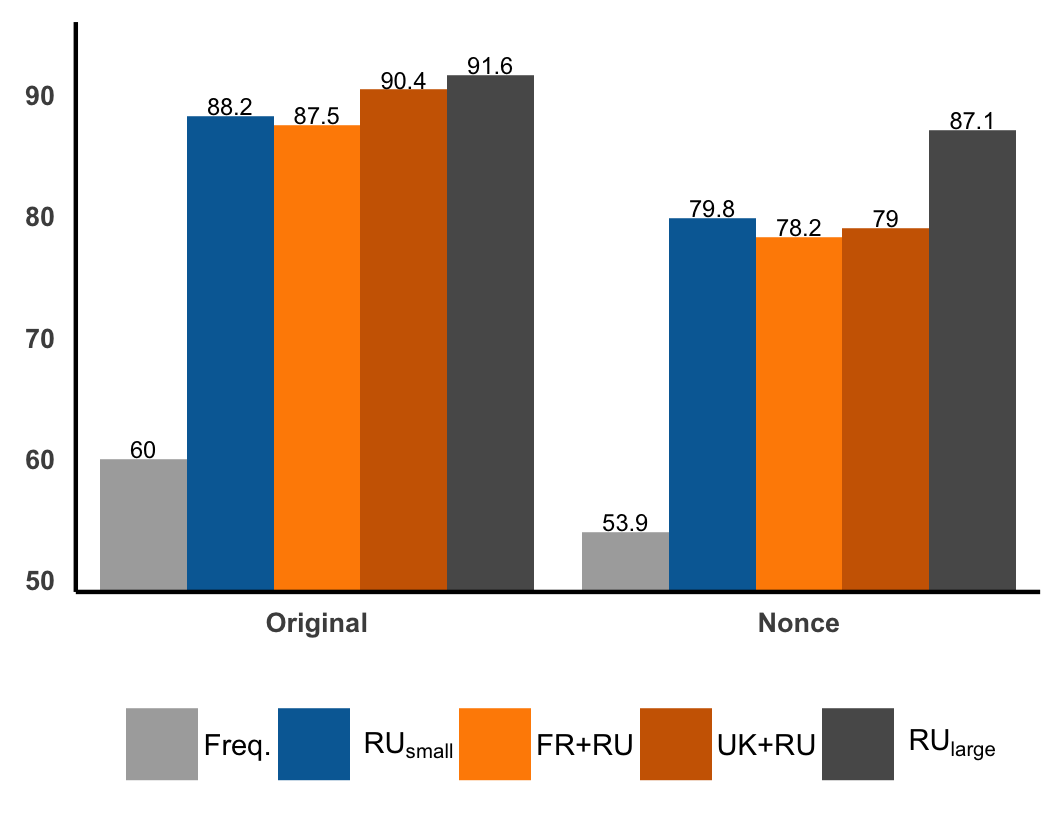}
\label{fig:ru}
\end{subfigure}
\caption{Probing Wikipedia-based monolingual and bilingual LMs on the agreement benchmark of \newcite{Gulordava2018}. 
Freq. is the Frequency baseline. Blue and black bars represent small and large L2 models, respectively. Orange bars represent bilingual models.}
\label{exp1}
\end{figure*}

\subsection{Main Results}
Figure \ref{exp1} shows the results. 
In this set of experiments, the bilingual models are trained by \textit{joint training} using the union of the vocabularies in the two languages (\textit{natural overlap}). See also Sect. \ref{agree}.
%
As in \cite{Gulordava2018}, the frequency baseline selects the most frequent word form (singular or plural) for each sentence. 

Looking at the Original sentences, we see that the bilingual models outperform the respective small monolingual models in the closely related pairs ES$\rightarrow$IT (86.8 \textit{vs} 79.8) and UK$\rightarrow$RU (90.4 \textit{vs} 88.2). 
However the addition of FR data results in lower accuracies on both $L2$s.
While this was expected in the unrelated pair FR$\rightarrow$RU, the large drop in FR$\rightarrow$IT is harder to explain.

When semantic cues are removed (Nonce sentences), ES$\rightarrow$IT is the only bilingual model to outperform its monolingual counterpart (80.7 \textit{vs} 79.4), while the accuracy drop in FR$\rightarrow$IT gets even larger (72.4 \textit{vs} 79.4).
This shows that exposing the model to a related language L1 is not guaranteed to improve implicit syntactic knowledge of L2, even when the rules of number agreement are largely shared between L1 and L2. On the contrary, our experiments suggest that in some cases L1 negatively interferes with the task in L2.




\subsection{Effect of Training Regime and Vocabulary Overlap on Agreement}
\label{sect:trainvocab}

Could transfer in FR$\rightarrow$IT be hampered by some of our experimental choices?
To consolidate our findings, we experiment with a different training regime (\textit{pre-training}) and a different vocabulary construction method (\textit{no-overlap}).
As shown in Table~\ref{tab:agreements}, both training regime and vocabulary overlap have a visible effect on the transfer of syntactic knowledge between FR and IT.
Pre-training considerably reduces the negative interference effect observed in joint training, and even leads to a higher accuracy on Original sentences in the no-overlap setup (83.2 \textit{vs} 79.8).
Eliminating vocabulary overlap (None) also leads to better agreement scores in most cases.
The best gain overall is obtained by the jointly trained model with no overlap (85.7 \textit{vs} 79.8) in the Original sentences, whereas no gain is observed in the Nonce sentences.

In summary, we find limited and inconsistent evidence of transfer of purely grammatical knowledge in our bilingual models.
Also contrary to our expectations, sharing more parameters (natural overlap) and mixing languages since the beginning of training leads to more negative interference than positive transfer in the FR-IT pair.

\begin{table}[ht]
\centering \small
\begin{tabular}{@{} c @{\ \ }c@{\ \ }c@{\ \ }c @{\ \ } c@{\ \ }c @{\ \ \ } c @{}}
& & \multicolumn{4}{c}{Bilingual (FR+IT$_{small}$)} & \\
\cline{3-6} 
& \multirow{2}{*}{IT$_{small}$} 
& \multicolumn{2}{c}{Joint Training}
& \multicolumn{2}{c}{ Pre-Training}
& \multirow{2}{*}{IT$_{large}$} \\
& & Natural & None & Natural  & None & \\
\midrule
Original & 79.8 & 74.8 & \textbf{85.7} & 79.8 & 83.2 & 86.6 \\
Nonce &79.4 & 72.4 & 77.6 & \textbf{77.7} & 76.8 & 79.4\\
\bottomrule
\end{tabular}
\caption{Impact of training regime and vocabulary overlap on agreement accuracy (FR$\rightarrow$IT).} 
\label{tab:agreements}
\end{table}

%

\begin{figure*}[ht]
\centering
\input{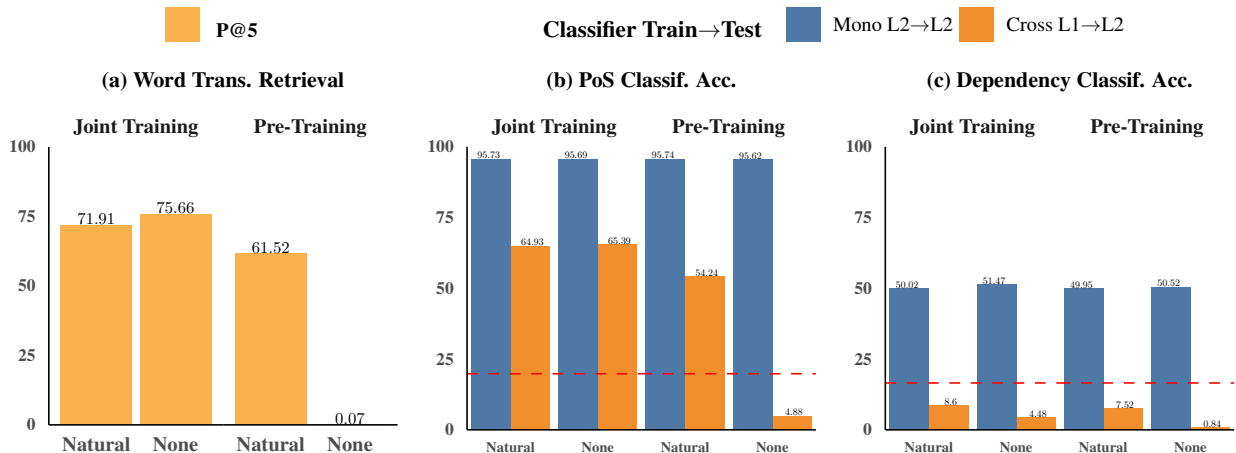}
\caption{Semantic \textit{vs} syntactic transfer in Wikipedia-based FR-IT bilingual LMs: (a) Word translation retrieval precision (P@5) measures lexical-semantic transfer; (b) PoS accuracy and (c) Dependency accuracy measure syntactic transfer. 
The classifiers are always tested on L2 (IT), and trained on either L2 or L1 (FR).
If syntactic categories were perfectly shared across languages, we should observe no drop between the blue and orange bars.
Dashed red lines show majority baselines for both (b) and (c).}
\label{rq1}
\end{figure*}

\section{Do mNNs Induce Shared Syntactic Categories?}
\label{sect:rq2}

Predicting long-range agreement is a rather complex task: in principle, besides learning agreement rules, the model has to discern several syntactic categories such as number, PoS and dependencies (e.g. distinguishing subject from other noun phrases). 
In practice, previous work \cite{ravfogel-etal-2018-lstm} showed that LSTMs sometimes resort to shallow heuristics when predicting agreement.

In this section we therefore investigate whether our mNNs induce at least basic syntactic categories that are shared across languages (RQ2).
We assume this is a necessary condition to enable transfer of purely grammatical knowledge, like agreement in nonce sentences, and beyond.

\subsection{Effect of Training Regime and Vocabulary Overlap on Syntactic Category Transfer}
\label{sect:trainingeffect}



In this section we examine the same FR-IT Wikipedia-based LMs described in section \ref{sect:trainvocab}.
Figure \ref{rq1}(a) shows that joint training yields better alignment of the word embedding spaces compared to the pre-training setup, which confirms the findings by \newcite{ormazabal19}.
Secondly, eliminating vocabulary overlap does not necessarily imply less alignment.
Interestingly, work on m-BERT/XLM models has also shown that vocabulary overlap has a much smaller effect on transfer than previously believed \cite{wu19emerging}.
An exception to this is the combination of pre-training and disjoint vocabulary (dubbed P/D), which gives near zero alignment of both lexical and syntactic spaces. This suggests that sharing hidden layers is not a sufficient ingredient to adapt a pre-trained model on a new (even if related) language, and that specific techniques should be used when joint training is not a viable option \cite{wang19-crossbert,artetxe19-crossbert}.

Moving to the transfer of syntactic categories (Fig. \ref{rq1}(b) we find that
all cross-lingually trained PoS classifiers (except P/D) perform much better than the majority baseline but notably worse than the corresponding monolingually trained classifiers.
As for dependency classification (Fig. \ref{rq1}c), accuracies are low overall and no 
cross-lingual classifier outperforms the majority baseline.
In summary, some form of syntactic transfer indeed occurs, but truly language-agnostic syntactic categories (such as \textit{noun} or \textit{subject}) have not emerged in our mNN representations.

\subsection{Training Objective, Number of Input Languages, and Language Relatedness}
\label{sect:rq2europarl}

We now study whether a different training objective, namely translation to a third language (English), leads to more syntactic transfer among input languages.
We also check whether number of input languages and language relatedness play a significant role in the sharing of syntactic categories.
All models in this section are jointly trained with natural vocabulary overlap on Europarl, and compared to their randomly initialized equivalents following \newcite{zhang-bowman-18}. Dependency classification results are omitted as they were always below the majority baseline. 

\begin{figure*}[ht]
\centering
\input{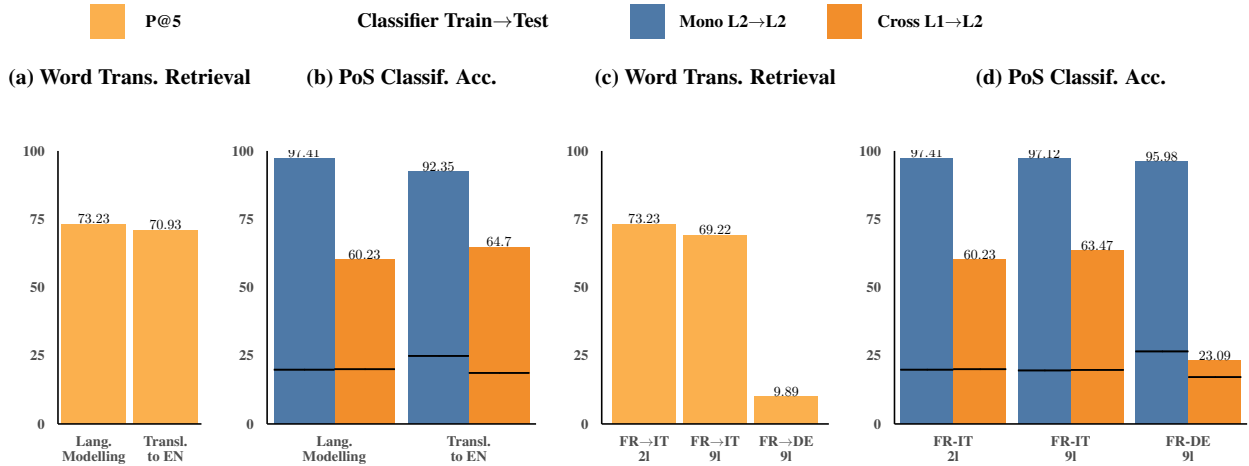}
\caption{Semantic (word translation retrieval) \textit{vs} syntactic (PoS classif.) transfer in Europarl-based bidirect. mNNs. 
(a,b) Effect of training objective: next word prediction \textit{vs} translation to English.
(c,d) Effect of number of input languages (2 \textit{vs} 9) and language relatedness (FR-IT \textit{vs} FR-DE) for the bidi-LM objective.
Horizontal lines (b,d) refer to the corresponding randomly initialized mNNs.
}
\label{fig:rq2}
\end{figure*}


\paragraph{Learning Objective}
As shown in Fig.~\ref{fig:rq2}(a,b), the translation objective has a slightly negative impact on the alignment of word embedding spaces when all other factors are fixed.
The translation objective also leads to lower PoS accuracy (monolingually probed), confirming previous results by \newcite{zhang-bowman-18}.
However, translating to English does result in visibly better cross-lingual transfer of PoS categories (mono/cross-lingual drop of $-27.7$ for translation \textit{vs} $-37.2$ for language modelling), showing that what are optimal conditions for lexical-semantic may no be optimal for syntactic transfer. 


\paragraph{Number of Source-side Languages}
For the remaining experiments we look at the (bidirectional) LM objective. 
As shown in Fig.~\ref{fig:rq2}(c,d), moving from 2 input languages to 9 results in lower WTR precision but higher cross-lingual PoS accuracy.
This suggests that adding more languages does not cause mNN representations to lose syntactic information and actually leads to more sharing of syntactic categories across languages. The generality of this remark is however restrained by our findings on language relatedness.

\begin{figure*}[ht]
\centering
\includegraphics[width=.8\linewidth]{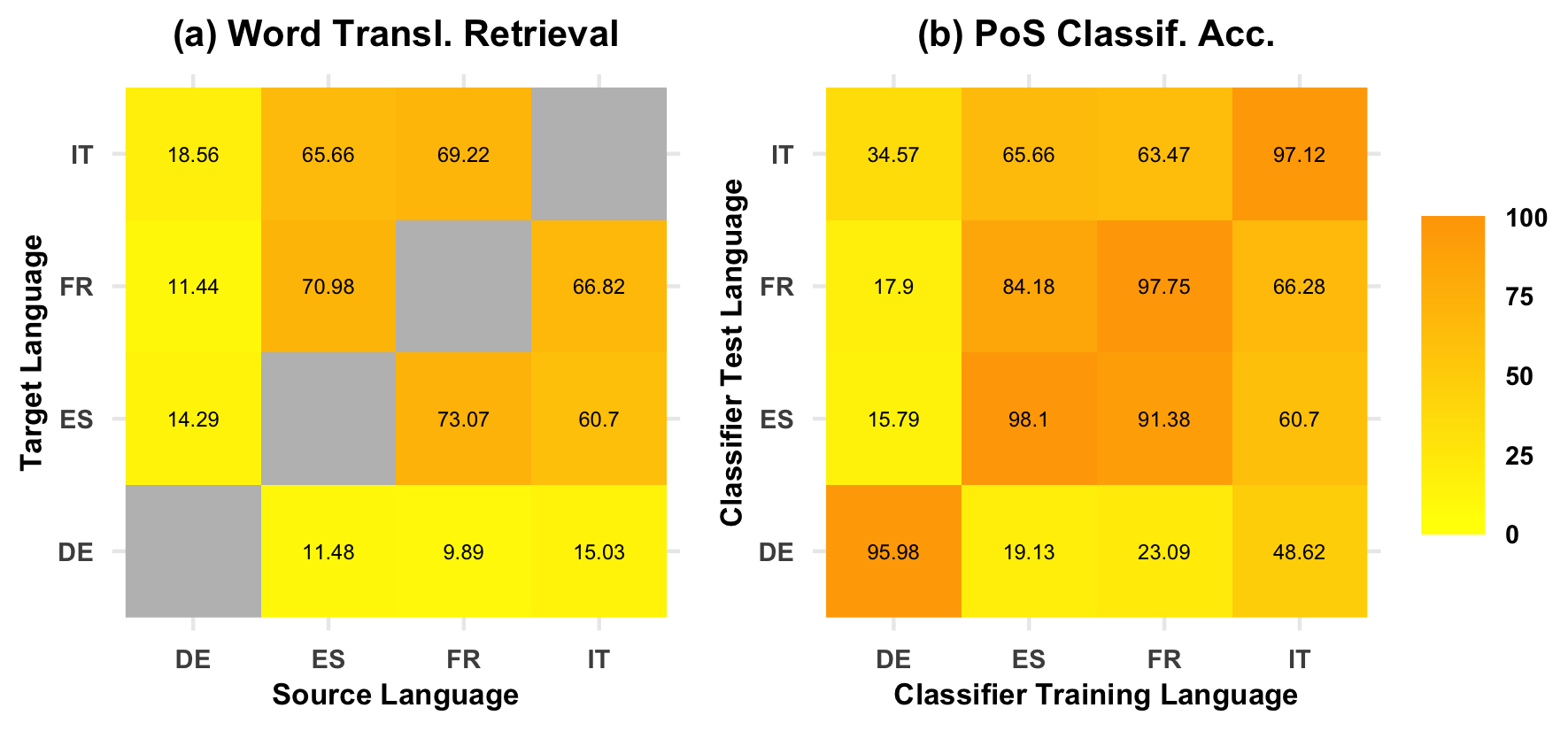}
\caption{Pairwise semantic and syntactic transfer in the 9-language bidi-LM (a subset of languages is shown).
Non-applicable (monolingual) settings in (a) are greyed out.
Diagonal values in (b) are scores of monoling. L2$\rightarrow$L2 classifiers, while remaining values are for cross-ling. L1$\rightarrow$L2 ones.}
\label{heatmap1}
\end{figure*}

\paragraph{Language Relatedness}
Fig.~\ref{fig:rq2}(c,d) also shows that moving from a very related pair of input languages (FR-IT) to a less related one (FR-DE) results in dramatically lower transfer of both lexical-semantics \textit{and} syntactic categories.
To substantiate this finding, we extend the analysis of our 9-language LM to more training-test pairs (we select a subset of languages for which a sizeable UD treebank exists).
The results in Fig.~\ref{heatmap1} confirm that, for both lexical-semantics \textit{and} syntax, the related languages FR, IT and ES report considerably higher values than those involving DE,
while the smallest drop ($-6.45$) is seen between FR$\rightarrow$FR and FR$\rightarrow$IT.
While we expected transfer to depend on relatedness, we did not expect the effect to be so large given that DE is not completely unrelated from the Romance languages.






\section{Conclusions}
We have presented an in-depth analysis of various factors affecting cross-lingual syntactic transfer within multilingually trained LSTM-based language (and translation) models.
Our main result is a negative one: Transfer of purely grammatical knowledge (specifically long-range agreement in nonce sentences) is very limited in general -- confirming recent findings by \newcite{mueller-etal-2020-cross} -- and strongly dependent on the specific choice of source-target languages. Namely, small gains were only reported on ES$\rightarrow$IT, while a considerable drop was reported on FR$\rightarrow$IT and almost no change was reported on UK$\rightarrow$RU.
When semantic cues were not removed (original sentences), transfer levels were overall higher with a peak of +7\% absolute in ES$\rightarrow$IT, but FR$\rightarrow$IT still suffered a considerable loss (-5\%).
While ES is arguably closer to IT than FR, we cannot yet find a convincing linguistic explanation for the large differences observed.
%
Our second set of experiments shows that POS categories are shared to a moderate extent, but dependency categories are not shared at all in our models. This suggests that syntactic knowledge transfer within our multilingual models is rather shallow, and may explain the lack of agreement transfer.

Our experiments with different training objectives and number of input languages show that what are optimal conditions for the alignment of word embedding spaces (lexical-semantic transfer) may not be optimal for syntactic transfer, and vice versa. 
Language relatedness is by far the most determining factor for both word embedding alignment and POS transfer. 
And finally, scaling from two languages to a mix of nine languages from three different families results in better POS transfer between related languages but considerably worse between unrelated ones.
Together with the findings by \newcite{wu19emerging}, our results suggest that scaling to highly multilingual models may improve syntactic transfer among the most related languages by decreasing the per-language capacity, but may also exacerbate the divergence among less related ones. Thus modern multilingual NNs appear still far from acquiring a true interlingua.

\section*{Acknowledgements}
Arianna Bisazza was partly funded by the Netherlands Organization for Scientific Research (NWO) under project number 639.021.646.


\bibliographystyle{acl_natbib}
\bibliography{nodalida2021}

\end{document}